\documentclass[journal,twoside,web]{IEEEtran}
\usepackage{cite}
\usepackage{amsmath,amssymb,amsfonts}
\usepackage{algorithm}
\usepackage{algpseudocode}
\usepackage{graphicx}
\usepackage{multirow}
\usepackage{xfrac}
\usepackage{math}
\usepackage{comment}

\usepackage{xspace}

\usepackage{review}
\setrevision{0}

\newcommand*{\eg}{e.g.\@\xspace}

\newcommand{\RRR}{{\mathbb R}}

\definecolor{green}{rgb}{0.0, 0.0, 0.0} 

\def\BibTeX{{\rm B\kern-.05em{\sc i\kern-.025em b}\kern-.08em
    T\kern-.1667em\lower.7ex\hbox{E}\kern-.125emX}}
    
\begin{document}
\title{Concept Graph Neural Networks for \\ Surgical Video Understanding}
\author{Yutong Ban, Jennifer A. Eckhoff, Thomas M. Ward, Daniel A. Hashimoto,\\ Ozanan R. Meireles, Daniela Rus, Guy Rosman
\thanks{Yutong Ban, Daniela Rus and Guy  Rosman are with Computer Science \& Artificial Intelligence Laboratory (CSAIL), MIT, Cambridge, MA 02139.  (e-mail: yban;rus;rosman@csail.mit.edu). }
\thanks{Yutong Ban, Jennifer A. Eckhoff, Thomas M. Ward, Daniel A. Hashimoto, Ozanan R. Meireles and Guy Rosman are with Surgical Artificial Intelligence and Innovation Laboratory (SAIIL), Massachusetts General Hospital, Boston, MA 02114, USA. (email: jeckhoff; tmward;  ozmeireles@mgh.harvard.edu)} 
\thanks{Daniel A. Hashimoto is with Department of Surgery, University of Pennsylvania Perelman School of Medicine, Philadelphia, PA USA. (email: daniel.hashimoto@uhhospitals.org)}
\thanks{The authors receive research support from the Department of Surgery at Massachusetts General Hospital as well as CRICO, NOSCAR and the Olympus Corporation.}
\thanks{$^{*}$corresponding email: yban@csail.mit.edu}
}

\maketitle

\begin{abstract}
 Analysis of relations between objects and comprehension of abstract concepts in the surgical video is important in AI-augmented surgery. Yet, building models that integrate our knowledge and understanding of surgery remains a challenging endeavor. In this paper, we propose a novel way to integrate conceptual knowledge into temporal analysis tasks using temporal concept graph networks. In the proposed networks, a knowledge graph is incorporated into the temporal video analysis of surgical notions, learning the meaning of concepts and relations as they apply to the data. We demonstrate results in surgical video data for tasks such as verification of the critical view of safety, estimation of the Parkland grading scale as well as recognizing instrument-action-tissue triplets. The results show that our method improves the recognition and detection of complex benchmarks as well as enables other analytic applications of interest.
\end{abstract}






\section{Introduction}


In many temporal analysis and prediction tasks, people leverage their conceptual understanding to make informed estimates \cite{edwards2010mind,teney2017graph}. Specifically, when reasoning about surgical workflow, surgeons leverage previously internalized concepts and understandings of the procedure and anatomy to support their decision making, foresee future operative steps, and comprehend the overall progress of the procedure.
In this paper, we focus on modeling different elements in surgery as concepts and try to analyze the relations between them. Examples for such underlying concepts include surgical instruments, organic tissues, surgical safety notions such as the "critical view of safety" (CVS) \cite{strasberg2010rationale}, and clinical variables, like the Parkland Grading Scale (PGS) \cite{madni2018parkland} for gallbladder inflammation, among other entities \cite{meireles2021sages,cameron2013current}. Correct modeling of the temporal evolution of different concepts and the relations of their underlying components is critical for the analysis of the surgery.   

\addnote[R2Q6]{1}{Computational analysis of surgical video data, however, often fails to generalize the model across different applications, as most of approaches focus on a specific task at hand (e.g. phase recognition or tool segmentation).} While achieving good results for specific procedures and machine learning (ML) tasks, such approaches do not easily adapt to different ML tasks and do not lend themselves to interpretation \cite{lipton2018mythos}. In addition, leveraging a structured environment allows ML models to better address problems for which we have an inductive bias 
\addnote[R3Q6]{1}{\cite{zambaldi2018deep,Lehrmann_undated-vz,Wehenkel2020-zw}}.
 Surgical decision-making is influenced by a large set of semantic concepts.
Correctly modelling these concepts and their relations offers a strong structural prior for data-driven models and a foundation for machines' comprehension of surgical workflow. \addnote[R2Q5]{1}{These concepts should be better investigated in order to improve computational support approaches, e.g. for surgical decision making.}
\begin{figure}
\centering
  \includegraphics[width=\linewidth]{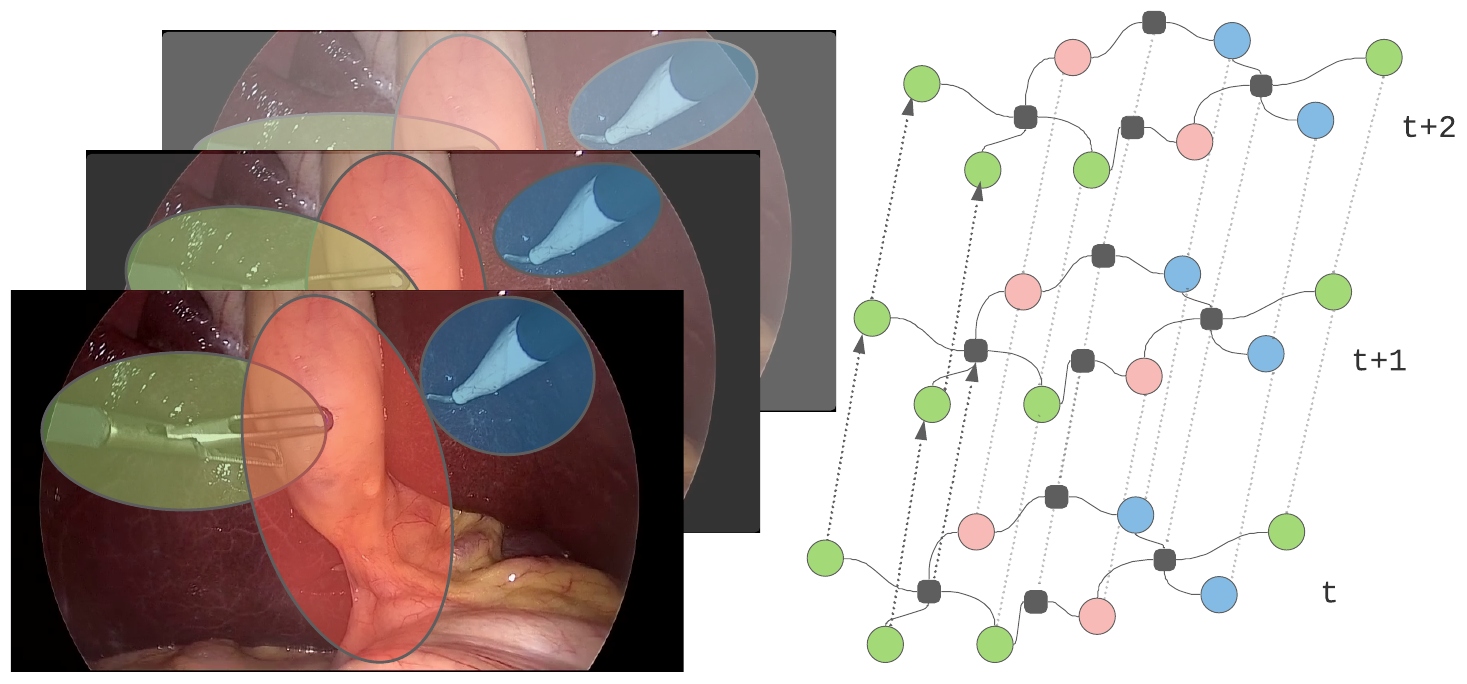}%
  \caption{Overview of the proposed model. The surgical process is modeled as a spatial-temporal graph. Elements (e.g. individual operative notions) in surgery are modeled as graph nodes (color points). The interactions of the elements are modeled as graph hyperedges (black squares) and inferred via message passing during video analysis. }
\end{figure}

Examples of how such concepts underlie surgical reasoning are prevalent in many procedures. However, in this paper we focus on a selection of clinically relevant tasks within laparoscopic cholecystectomy. Due to its highly standardized procedural steps, its stable field of view, and high volume numbers across the world, laparoscopic cholecystectomy lends itself to machine learning purposes and presents a benchmark in surgical AI. 
The conceptual surgical notions that afford structured reasoning about causes of operational risks in laparoscopic cholecystectomy manifest in several examples. The critical view of safety (CVS) presents one of these safety measures. CVS is defined as the clear dissection and visualization of the cystic duct and cystic artery, the clearing of the hepato-cystic triangle of tissue and exposure of the cystic plate \cite{vettoretto2011critical}\cite{strasberg2010rationale}. An additional surgical notion, the Parkland grading scale (PGS), provides a perspective on the degree of inflammation of the gallbladder and the severity of the underlying disease, upon initial inspection \cite{madni2018parkland}. PGS involves assessing the degree of adhesions, hyperemia, gallbladder distension, intra-abdominal fluid, liver abnormalities and necrosis or perforation of the gallbladder \cite{madni2018parkland}. Higher PGS grades are associated with longer operative times, higher complication rates and elevated morbidity and mortality, making PGS a highly relevant concept to assess during laparoscopic cholecystectomy. Furthermore, the interaction between instruments, performed actions and tissues, commonly referred to as "action triplets", provides a granular understanding surgical action and consequence, allowing for fundamental comprehension of surgical workflow. These concepts are all defined as a set of characteristics that relate a set of observable factors. Enabling machines to comprehend surgical workflow and assess these relevant concepts may significantly contribute to the application of machine learning in the operating room, facilitating intra-operative decision making and risk mitigation, as well as holding great teaching potential for surgical trainees. 

{\begin{figure*}[h!]
\centering
\includegraphics[width=1.0\linewidth]{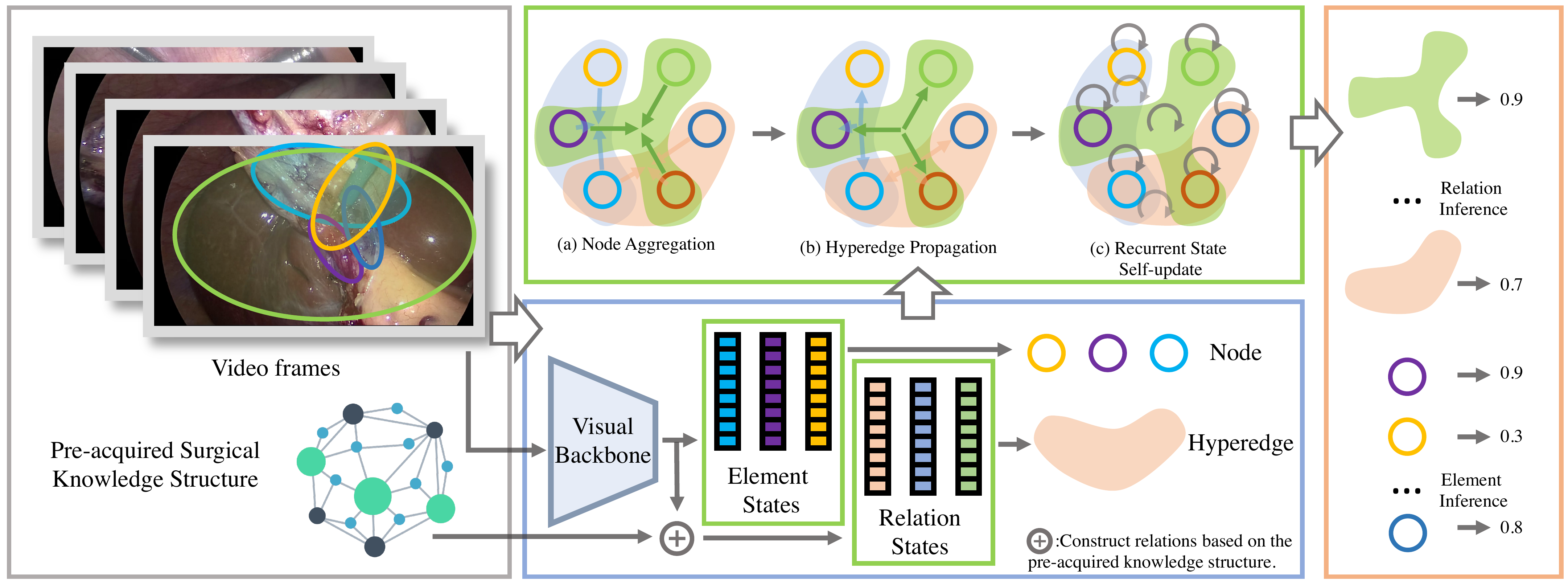}
\caption{\color{green}{Overview of the proposed concept graph model. After a visual backbone to extract visual features, we use the pre-acquired surgical knowledge structures to form the relations to inject problem inductive bias into the model. The elements and relations are then represented as nodes and hyper-edges in the graph. At each time $t$, the graph performs three different steps (a) Node aggregation (b) Hyperedge propagation and (c) Recurrent state self-update. After the three steps. the model emits the probability of presence for each element/relation.}}
\end{figure*}}
In this work we propose a model that mimics the structure of the problem and explores temporal reasoning capability. Our approach leverages a knowledge hypergraph model, suitable for learning, making inferences and predicting different concepts in surgical video streams. The approach can use different amounts of supervision for different aspects of the problem, as well as offers explainability of network structures and their outputs. Furthermore it may easily adapt to various surgical tasks and procedures in the future, hence promising in tremendous clinical potential.

\addnote[R2Q2]{1}{The main contributions of the paper are: 
\begin{itemize}
    \item We present a novel knowledge hypergraph framework ConceptNet for online temporal inference of the entities and the relations in surgical videos.
    \item We show how ConceptNet framework allows us to translate the prior clinical problem description into a graph-based inductive bias for the problem. We show a set of steps to convert the problem description into the graph topology for the network.
    \item We experimentally prove that under data constraints, the proposed model and associated graph topologies have a large performance benefit by incorporating the problem structure into the model. The surgical problems tackled in the paper include two novel constructed datasets that relate to safety-related practices and risk estimation in surgery.
\end{itemize}
}

\section{Related works}
Our approach relates to several active fields of research. In surgical computer vision, current efforts are focusing on analysis of temporal comprehension of surgery, such as  operative phase recognition for laparoscopic and endoscopic surgeries \cite{ban2021aggregating, volkov2017machine, ward2020automated} or prediction of remaining duration of the procedure \cite{twinanda2018rsdnet}, as well as spatial segmentation and detection of important surgical entities such as tools and the related actions \cite{twinanda2016endonet, jin2017sv, nwoye2020recognition}.  We demonstrate effectiveness of our approach on laparoscopic cholecystectomy, as it has been frequently focused on in  computer vision analysis and ML applications, due to its relatively standardized workflow
\cite{jin2017sv,twinanda2016endonet,twinanda2018rsdnet,jalal2019predicting,manabe2019cnn,ban2021aggregating,ban2021surgical},
as well as conceptual notions such as CVS, as a safety measure for prevention of accidental clipping of the common bile duct, and PGS, as an objective estimation of gallbladder inflammation allowing for inference on the severity of the disease.  \cite{nwoye2020recognition,namazi2020ai,mascagni2021artificial,ward2022artificial}.

As our model is a representation of the various surgical notions as the states of elements of a knowledge hypergraph estimated over time, inspired by the work in the knowledge graphs community. Graphical knowledge representation has been extensively studied in the literature \cite{nickel2015review,fatemi_knowledge_2020} and more recently in a growing community centered around graph neural networks \cite{wu2020comprehensive,battaglia_relational_2018,wang2019knowledge,feng2019hypergraph,deng2020dynamic}. In contrast to the proposed knowledge graph for surgical videos, most of the existing works in the field involve static information, rather than relying on sensor data streams or inferring from and predicting on the basis of these streams. Furthermore, the majority of knowledge graph literature merely utilizes pairwise node relations for each edge. Indeed, the pairwise node relation affords simple coding and allows multi-concept relations via a set of edges and nodes \cite{fatemi_knowledge_2020}. However, in applications with more complex problem structures, it may require additional encoding which can complicate the representation and result in loss of the explainability.

Graph neural networks (GNNs) \cite{scarselli2008graph,bronstein2017geometric,battaglia_relational_2018,wu2020comprehensive} have recently been applied in many computer vision tasks, \addnote[R1Q2]{1}{such as semantic segmentation \cite{chen2019graph}, intent prediction \cite{liu2020spatiotemporal}, skeleton based action recognition \cite{li2019actional, shi2019skeleton}}
and shown their advantage of modeling spatial and temporal aspects of the problem. 
However, the nodes in these graphs are often describing either physical entities or hard-coded attributes that are tailored to the ML task and algorithmic approach. Some approaches have been more general, and yet they are often explicitly representing visible entities in each image 
\cite{sabour2017dynamic,Choi_2019_ICCV}.
When reasoning about multi-agent behavior, GNNs are extensively used to model pairwise interactions 
\cite{salzmann2020trajectron++, vemula2018social},
with the majority of the approaches connecting graphs at few time points due to the lengths of the sequences involved.
The modularity and generality of graph neural networks has been advantageous for representing complex relationships and inter-dependency between objects in a variety of other applications \cite{kipf2019contrastive, ermolov2020latent, liu2017skeleton, cai2019exploiting}. Such advantages are fitting for surgical video analysis problems, as surgical relations and interactions are highly structured.


\vspace{-2ex}
\section{Model}
\subsection{Problem Formulation - Structured Surgical Analysis}
\color{black}

Understanding surgical workflow from a minimally invasive video can be formulated as two complementary processes: (i) inferring the state of relevant surgical concepts; (ii) reasoning the relations among those concepts. 

Therefore, we define the problem of structured surgical data analysis as follows: given a video sequence with image frames $\left\{I^{t}\right\}_{t=1}^{T}$, (i) infer the state of surgery in terms of a set of $N$ concepts that pertain to the surgery at each time frame at time $t$, $I^{t}$. Concepts in videos can include visible objects, or latent, higher-level abstract notions deduced from the surgery. Inferring their latent state can be expressed as the inference of a set of emitted outputs about each concept, e.g. the visibility of the gallbladder in a specific frame, or if a particular operative action is taking place at that time. (ii) Aside from the individual states of the concepts, we also infer emitted labels regarding $M$ relations that relate the concepts, given as a set of outputs based on the state of each relation.

\vspace{-2ex}
\subsection{Surgical Concept Graph Networks}
In this section, we define the Surgical Concept Graph Network (ConceptNet). Inspired by how surgeons use conceptual knowledge to understand and interpret surgical video data,
we use a knowledge graph \cite{nickel2015review} for laparoscopic surgical videos, applying a hypergraph structure to capture conceptual surgical knowledge and the relations between these concepts. 
For instance, if there is a clinical phenomenon that involves several individual factors, the factor of the phenomenon being individually achieved and composing the whole phenomenon achieved can be modeled as an evolution of a dynamic graph. In addition, as relations in surgery often involve more than two elements, we extend the standard graph neural network with hypergraphs \cite{bretto:hal-01024351}. We use the standardized notation for the graph neural network, as a basis notation of our network. A hypergraph neural network is formally written as: 
\begin{equation}
    H=(V,E) 
\end{equation}
where $V$ is the set of nodes and $E$ is a set of hyperedges that connect the nodes. Unlike pair-wise graph edges that capture binary relations, hyperedges capture more general \textbf{n-nary} relations, and are able to connect more than two elements, making it possible to directly model the interactions among multiple elements in surgery, and do so efficiently \cite{feng2019hypergraph,fatemi_knowledge_2020}.

\noindent The components of the proposed model are defined as follows:\\
\noindent \textbf{Nodes}: Let $n \in \{1, ..., N\}$ denote the index of concepts in a surgery. At each time step $t$, a node on the surgical graph is represented by a vector $v_n^t \in \RRR^{d_n}$. Example nodes include the achievement of full dissection of the individual components of CVS, as mentioned in the introductory section of this paper, such as "cystic artery" and "cystic plate", or different surgical instruments appearing in a surgical video. Each node is modeled as an LSTM state. We use a shared-weight node LSTM module to propagate the state of each concept over time. The input of the LSTM is derived from both input encoders of the image frames, and from the aggregation of messages from the connected edges (see Fig. \ref{fig:message_passing}).

\noindent \textbf{Hyperedges}: Since each hyperedge involves explicit semantics, we choose to use explicit state to model each hyperedge. Each hyperedge state is represented by the vector $e_k^t \in \RRR^{d_e}$. 
Example hyperedges can describe the relationship between CVS and its components or the definition of a PGS as described by possible indications. In practice, each edge state is also processed by a shared-weight edge LSTM. The input of the edge LSTM consists of the visual features of the visual backbone and the aggregated features of the connected nodes.  

\noindent \textbf{Global Identity Vector}: We instantiate an identity vector for each node and hyperedge, and feed it as an input into the temporal node/hyperedge encoder. This allows us to describe the semantics of each node and hyperedge across the entire dataset.

Overall, our model forms a semantically-structured GNN inference, that is, encoding the graph state based on temporal video input. We further extend the semantically structured GNN to temporal inference by conditioning message passing operations between nodes, edges, and the graph upon encoding of per-frame video input $I^{t}$. This approach models the analysis of the temporal surgical process, updating our posterior beliefs about the entities and relations in the graph. Updates are made by interleaving the node-to-hyperedge and hyperedge-to-node message passing steps \cite{kearnes2016molecular,battaglia_relational_2018} between frame update steps, as we show in Section~\ref{subsec:message}.

\vspace{-2ex}
\subsection{Message Passing}
\label{subsec:message}
\begin{figure}

\centering
  \includegraphics[width=1.0\linewidth]{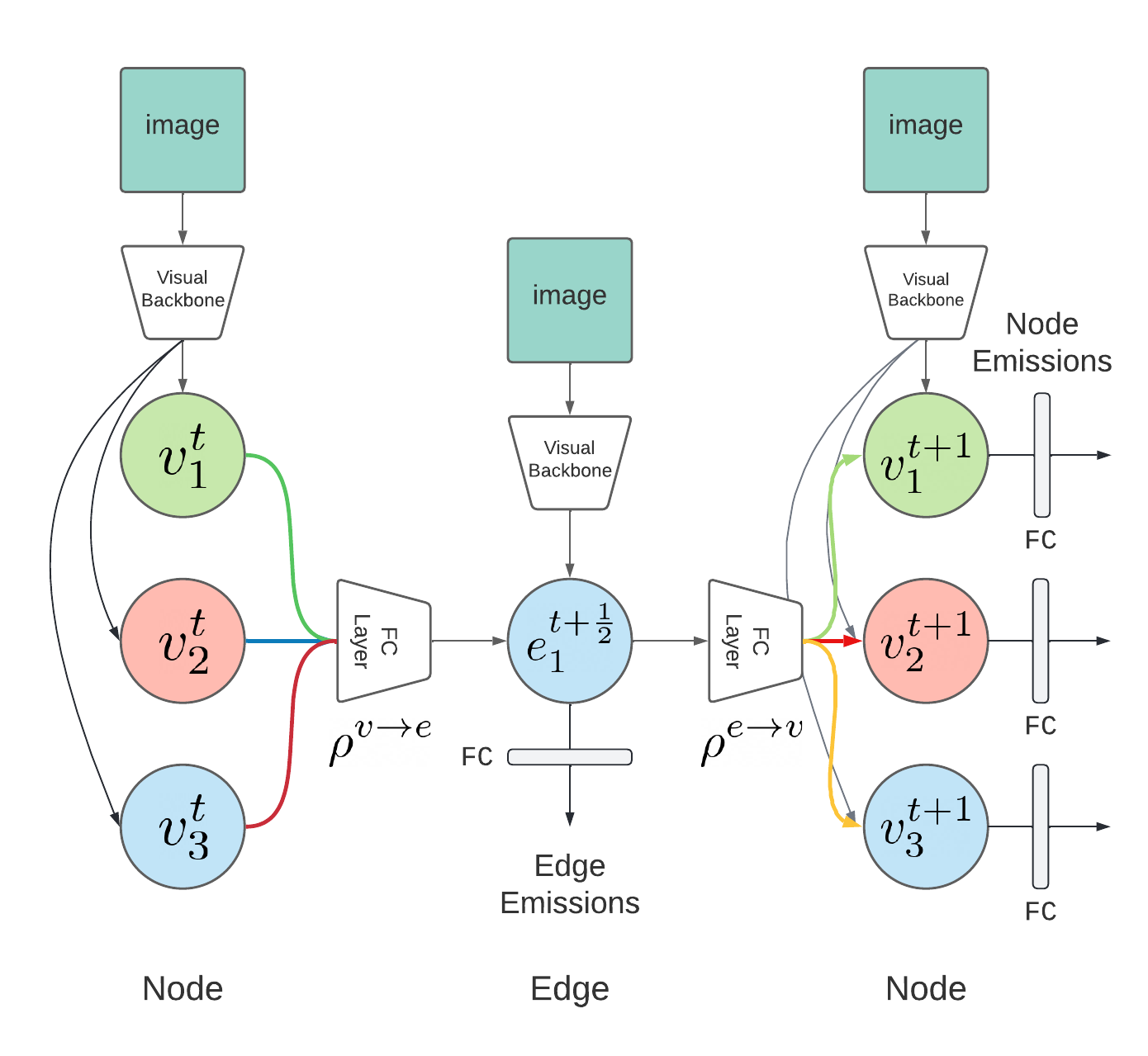}%
  \caption{Overview of the message passing within our model. Each circle represents an LSTM. The message was passed from node-to-edge then edge-to-node. Each message passing step is condition on an encoding of the image.}
  \label{fig:message_passing}
\end{figure}

\addnote[R3Q4a]{1}{
Message passing models the communication between the different node/edge states, which in practice involves three steps, Node aggregation, Hyperedge Propagation and Recurrent state self-update.
\subsubsection{Node aggregation} While our model is in the message passing iterations, the input image encoding is also involved. For each frame we first compute an \emph{edge update}, to aggregate the node information to the related hyperedge, formally written as:
\begin{align}
   \label{eq_node_agg}
\bar{v}_k^t &= \rho^{v \rightarrow e}({V}_k^{t}), \ {V}_k^{t} = \{v_i^{t}\}_{i\ s.t.\  k \in \mathcal{N}_i}\\
   e_k^{t+\half} &= \phi^{E}(e_k^{t},\bar{v}_k^t,u^{t},I^{t}),
   \label{eq_edge_update}
\end{align}
where $v_i^{t}$ denote $i{\text-}th$ node state at time $t$, and $e_k^{t}$ denotes $k{\text-}th$ edge states at time $t$, $\mathcal{N}_i$ denotes the neighborhood of node $i$ and ${V}_k$ denotes the node set that connects to edge $k$. We also have $\rho^{v\rightarrow e}$ representing node-to-edge aggregation operators and $\phi^{E}$ defining node-to-edge neural message passing updates. And $u^{t}$ is the global concept vector. We use $t+\half$ to denote intermediate steps as in temporal numerical schemes \cite{morton2005numerical}.
\subsubsection{Hyperedge probagation} 
The edge update is followed by an edge-to-node \emph{aggregation} step,
\begin{align}
   \label{eq_edge_agg}
\bar{e}_i^{t+\half} &=\rho^{e \rightarrow v}({E}_i^{t+\half}),\ {E}_i^{t+\half} = \{e_k^{t+\half}\}_{k \in \mathcal{N}_i}
 \\ 
v_i^{t+1}&= \phi^{V}(\bar{e}_i^{t+\half},v_i^{t},u^{t},I^{t}) 
\label{eq_node_update}
\end{align}
where ${E}_i$ denotes the edges set that connects to node $i$.  $\phi^{V}$ defines edge-to-node neural message passing updates, and $\rho^{e\rightarrow v}$ denotes aggregation operators based on the inclusion of nodes in hyperedges.
\subsubsection{Recurrent state self-update} 
Each concept node and each hyperedge are represented by an LSTM temporal module that encodes their current state. The temporal module fuses incoming messages in addition to the current image encoder and the current LSTM state, written as:
\begin{align}
  \label{eq_temporal_update}
  \phi^{E}(e_k^{t}, &\bar{v}_k^t,u^{t},I^{t}) = \nonumber \\ & LSTM(h_e^{k,t}, f_e(h_e^{k,t}, e_k^{t},\bar{v}_k^t,u^{t},I^{t}))&&\\
      \nonumber
  \phi^{V}(\bar{e}_i^{t+1/2}, &v_i^{t},u^{t},I^{t}) = \\ &  LSTM(h_v^{i,t}, f_v(h_v^{i,t},\bar{e}_i^{t+1/2},v_i^{t},u^{t},I^{t}))&& 
\end{align}
where $f_{e}(\cdots),f_{v}(\cdots)$ are encoders of both the other neighboring nodes and hyperedges, as well as the current 
\color{green}image. See Section~\ref{sec:graph_design} for a more detailed discussion and several architectures we have used.
We use dropouts of entire nodes and edges as we perform the message passing, regularizing the network as often done in structured neural networks, \eg in \cite{velivckovic2018graph,huang2019uncertainty,do2021graph}. The message-passing process is illustrated in Fig.\ref{fig:message_passing}.
The algorithm is written as Algorithm~\ref{alg_concepts}. 
}
\color{black}

\emph{Directed graph networks} Our model can incorporate both directed and undirected hypergraphs. Directed hyperedges are represented as $(v_{in},v_{out})$, following the notation in \cite{bretto:hal-01024351}, where $v_{in},v_{out}$ are the incoming and outgoing nodes respectively. Directed edges allow us to imply a chain of reasoning \cite{koller2009probabilistic} or causal assumptions as we describe thought processes and neural inference \cite{koller2009probabilistic}. In practice, the clinical applications on which we focus have a clear and directed formulation, as we show in Section~\ref{sec:graph_design}.


\begin{algorithm}
\begin{algorithmic}
\Procedure{GraphUpdate}{$e^t,v^t$}
  \State Update node aggregations using Equation~\ref{eq_node_agg}. This includes global identity vector updates.
  \State Update edge updates using Equation~\ref{eq_edge_update}.
  \State Update edge aggregations using Equation~\ref{eq_edge_agg}.
  \State Update node updates using Equation~\ref{eq_node_update}.
  \State Generate spatial/temporal emissions based on node and edge temporal models using Equation~\ref{eq_temporal_update}
  \EndProcedure
\end{algorithmic}
\caption{Temporal processing with concepts networks. \label{alg_concepts}}
\end{algorithm}

\begin{figure*}[t!]
\vspace{-4ex}
\centering
\includegraphics[width=.99\textwidth]{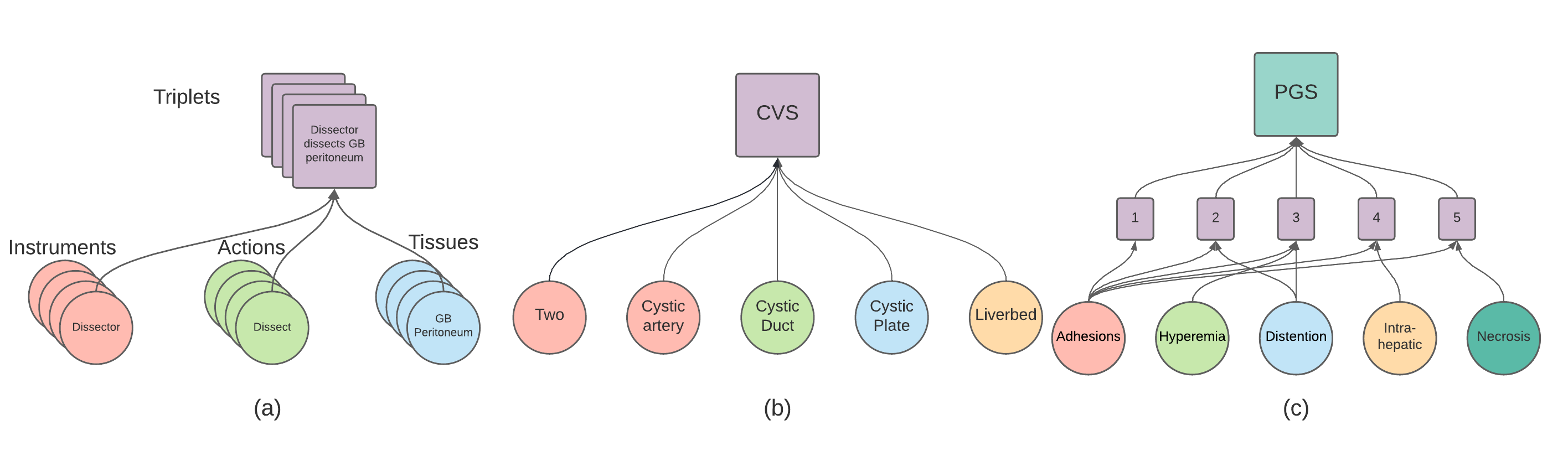}
\vspace{-4ex}
\caption{The graph structures translated from different clinical problems. (a) Action Triplet Recognition (b) CVS (c) PGS Recognition. The colored circles indicate the graph nodes and the squares indicate the graph hyperedges.}

\label{fig:graph_design}
\end{figure*}

\vspace{-2ex}

\subsection{Image Encoding}
\label{subsec:image_encoding}

Several architectures can be used for the image encoding and message-passing functions $f_e,f_v$, and the ConceptNet approach integrates seamlessly with how these architectures would traditionally perform per-image classification tasks. 

\color{green}
 {
A straightforward approach is to use image features output from a visual backbone. The output image feature  is then concatenated along with the global vector and the aggregated neighboring nodes, to provide input to each concept LSTM. In doing so, our approach resembles a }
\color{black}

\noindent  {\color{green}compound version of a spatial-temporal network, which has been shown to be suitable for surgical analysis tasks in the past \cite{twinanda2016endonet,hashimoto2019computer}. During training, a random dropout of the image feature/aggregated neighbor vector is applied, which forces the network to learn from the individual modal of the information, either from the image feature or the connected hyperedge information. }
\color{black}

\addnote[R2Q4a]{1}{ 
The concepts network is compatible with any visual backbones that can extract the image features. For convolutional networks, such as ResNet\cite{he2016deep}. we can use convolutional layers and early fully-connected layers as feature extractors.  For more recent multi-head-attention-based backbones like vision transformers \cite{dosovitskiy2020image}, we use the output of \texttt{[CLS]} token as the image feature representation, akin to the image classification task. We then feed the corresponding feature into the message-passing function of either the node-to-edge or edge-to-node steps.
}


\vspace{-2ex}

\subsection{Network Emissions and Annotations}
The network is not limited to emitting the existence of the concepts per frame. In practice, emissions may include various possible surgical annotations: i) the existence of concepts with different temporal predicates ("Is the liver present in the image?", "Is there bleeding?") ii) Spatial information about concepts ("What is the position of a cystic duct within the image?"). iii) Additional information about concepts ("What is the pose of a clipping tool?"). Throughout this primary exploration of the possible applications in surgery, we focus on temporal annotations, indicating the presence of a particular concept in the image, which illustrated that certain components of the concept had been accomplished. The emission network for each concept is one fully connected layer, which projects the LSTM hidden state to the emission probabilities.


\vspace{-2ex}
\addnote[R3Q4b]{1}{
\section{{Surgical Graphs for Specific Tasks} \color{black}
\label{sec:graph_design}
}
}

In this section, we show the detailed graph design for each clinical ML task. The graphs for different problems are illustrated in Fig \ref{fig:graph_design}.
\addnote[R2Q2b]{1}
{
The construction of the graph follows a general way of using the pre-acquired surgical knowledge structure for the graph design. For leveraging clinical knowledge as an inductive bias in our framework, for each of the specific problems, we take the following steps:
\begin{itemize}
    \item We extract from the clinical description the relevant notions, and concepts that are pertinent to the problem description.
    \item We extract the logical-temporal relations in the problem descriptions, and form hyperedges to match them.
    \item We partition the participating nodes in each hyperedge according to an input-output relation in the case according to the clinical definitions. In the case that the relation is there, but there is no clear functional dependency, we can mark all of the participating nodes as inputs, and let the neural networks infer the semantic relation (effectively creating a factor in a neural factor graph).
\end{itemize}
We apply this approach to form a model for three clinical video analysis tasks as we show below.
}
 
\vspace{-2ex}
\subsection{Surgical Action Triplet Recognition}
The first task we are focusing on is the action-triplets-recognition task\cite{innocent2021rendezvous} in laparoscopic cholecystectomy. In this task, the main surgical elements that compose an operative procedure are categorized into three groups: instruments, actions and tissues. Moreover, the relations between the elements are interconnected and labeled as triplets, e.g. ${\langle\text{dissector},\text{dissect},\text{gallbladder}\rangle}$ is a triplet. Our goal is to correctly identify the presence of these surgical elements as well as the interaction among them. In the action triplet recognition task, each element is represented as a node on a graph. At each time step, a node emits a binary estimate that indicates the presence of the element in the current frame. Every possible triplet combination is represented by a hyper-edge. Similar to the node emissions, each triplet estimation is emitted by the hyper-edge.   

\begin{table}[]
    \caption{Labels included in the CVS100 Dataset}
    \centering
    \begin{tabular}{c|c|c}
    \hline
    Class & Description & Per-frame labels \\
    \hline
    CVS&CVS is achieved &  yes, no \\
    Cystic plate &Cystic plate is exposed &  yes, no\\
    \multirow{2}{3em}{Two} &Two and only two structures   &  \multirow{2}{3.2em}{yes, no } \\
    &are dissected and exposed & \\
    Cystic artery&Cystic artery is exposed & yes, no \\
    Cystic duct&Cystic duct is exposed & yes, no \\
    \hline
    \end{tabular}

    \label{tab:cvs_dataset}
\end{table}

\begin{table}[]
    \caption{Labels in the ParklandGradingScale200 Dataset}
    \centering
    \begin{tabular}{c|c}
    \hline
    Class & Labels \\
    \hline
    PGS & 1,2,3,4,5\\
    Adhesion & body, buried, majority, neck, none\\
    Distention & distended, normal, shrivelled\\
    Hyperemic & yes, no \\
    Intra-hepatic & yes, no \\
    Necrotic & yes, no \\
    \hline
    \end{tabular}

    \label{tab:pgs_dataset}
\end{table}

\vspace{-2ex}
\subsection{Critical View of Safety}
Another clinical notion we address is the critical view of safety (CVS) in laparoscopic cholecystectomy.  CVS is defined as the clear dissection of several anatomic landmarks: the cystic duct, the cystic artery, the cystic plate, and the hepatocystic triangle \cite{felli2019feasibility}. It is a systematic safety measure designed to prevent misidentification of anatomic structures and subsequent accidental injury of the common bile duct and associated injury of the hepatic artery. Common bile duct injury presents a major complication of laparoscopic cholecystectomy and is associated with an immense increase of morbidity and mortality rates\cite{renz2017bile}.  The full achievement of CVS can be challenging, especially with increasing severity of gallbladder pathology that leads to major adhesions and tissue fragility. 
Specifically in cases where CVS is hard to achieve, automated recognition would augment surgical safety and provide additional supervisory clues to the surgeon. Within this concept model, the individual components of CVS are modeled as nodes and the overall achievement of CVS is defined as a hyperedge.

\begin{figure*}[ht!]
\vspace{-5ex}
\centering
\begin{tabular}{c}
\includegraphics[width=0.98\textwidth]{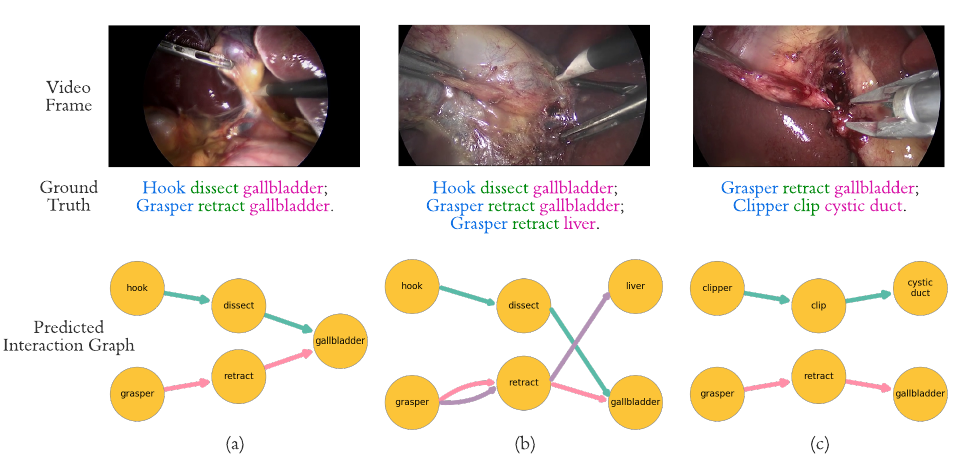}
\end{tabular}
\caption{\color{green} {Example results in CholecT45 validation dataset. (a) is the last frame of the sequence. (b) indicates the ground truth activations. (c) is the estimated graph activation. In both (b) and (c), only the activated nodes are plotted, links between the nodes represent the joint activation of the action triplet.}}
\label{fig:results_triplet}
\vspace{-2ex}
\end{figure*}

\begin{table*}[h]
    \caption{Results on CholecT45 Dataset. Average precision (AP) is shown for different components and different combinations of components.  The proposed ConceptNet achieved the best $AP_{IVT}$ by better modeling of the visual-temporal structure between the components.}
    \centering
    \begin{tabular}{c|ccc|ccc}
    \hline
    Method & $AP_{I}$ & $AP_{V}$ & $AP_{T}$ & $AP_{IV}$ & $AP_{IT}$ & $AP_{IVT}$ \\
    \hline
    Tripnet\cite{nwoye2020recognition} (MICCAI 20) & $\textbf{89.9} \pm \textbf{1.0}$ & $59.9 \pm 0.9$ & $37.4 \pm 1.5$ & $31.8 \pm 4.1$ & $27.1 \pm 2.8$ & $24.4 \pm 4.7$ \\
    Attention Tripnet \cite{innocent2021rendezvous} (MIA 22) & $89.1 \pm 2.1$ & $61.2 \pm 0.6$ & $40.3 \pm 1.2$ & $33.0 \pm 2.9$ & $29.4 \pm 1.2$ & $27.2 \pm 2.7$\\
    Rendezvous \cite{innocent2021rendezvous} (MIA 22) & $89.3 \pm 2.1$ & $62.0 \pm 1.3$ & $40.0 \pm 1.4$ & $34.0 \pm 3.3$ & $30.8 \pm 2.1$ & $29.4 \pm 2.8$ \\ 
    \hline
    $\textbf{ConceptNet-Resnet50 (Ours)}$  &  $84.1 \pm 2.9  $&  $58.7\pm 1.3 $ & $ 42.4 \pm 2.3 $ & $\textbf{38.6} \pm \textbf{2.0}$ & $29.8 \pm 1.7$  & $ 30.1 \pm 2.8$\\

    $\textbf{ConceptNet-VIT (Ours)}$ & $88.3 \pm 1.4$ & $\textbf{65.1} \pm \textbf{3.9}$& $\textbf{43.8} \pm \textbf{3.1} $ & $33.7 \pm 3.2$& $\textbf{33.4} \pm \textbf{3.2}$& $\textbf{30.6} \pm \textbf{1.9}$\\
    \hline
    \end{tabular}
    \label{tab:results_triplet}
\end{table*}

\vspace{-2ex}
\subsection{Parkland Grading Scale}
The Parkland grading scale (PGS) provides information about the degree of gallbladder inflammation upon initial inspection. Among other intraoperative grading scales, it assesses gallbladder inflammation into 5 categories, according to gallbladder distension and anatomy, amount of adhesions, hyperemia, necrosis and perforation \cite{madni2018parkland}.
Previous work has demonstrated a strong correlation between higher PGS, hence more severe inflammation, with longer operative duration, higher conversion rates to open surgery, and more frequent complications, such as bleeding, postoperative bile leakage, and readmission \cite{sugrue2019intra, griffiths2019utilisation}.  \addnote[R2Q3c]{1}{The prior problem structure of the PGS follows its clinical definitions in \cite{madni2018parkland}. To translate the clinical definition in our model, the factors of the PGS are represented as nodes and the scale itself is a relation that connects the necessary components, as illustrated in Fig. \ref{fig:graph_design} (c).} 


\section{Experiments}
In this section, we introduce the surgery video dataset, the detailed experimental settings, the evaluation metric, as well as the results of each specific task. 
\label{sec:results}
\vspace{-3ex}
\subsection{Dataset}
\label{sec:dataset}
\subsubsection{CholecT45 Dataset} The CholecT45 Dataset \cite{innocent2021rendezvous}, a subset of \cite{nwoye2022cholectriplet2021} is used for detailed, granular surgical understanding. Three different sets of labels are provided for each video frame, representing 6 surgical instruments, 10 actions, and 15 tissues. Overall, the annotations provided of the surgical components result in 100 triplets of ${\langle \textit{instrument,action,tissue}\rangle}$ interactions. In the dataset, videos are down-sampled into 1 fps, yielding in total of 45 surgical videos, 5 different splits were used to cross-validate the model. 

\subsubsection{ParklandGradingScale200} The dataset that was used contains 200 videos of the initial inspection phase. This inspection phase takes place immediately after port placement and serves as an period of observation of the circumstances (e.g. general anatomy, presence of pathological adhesions, degree of gallbladder inflammation). The inspection phase is followed by the retraction of the gallbladder fundus. \cite{ban2021aggregating}. Each inspection phase within the dataset is 16 seconds long. The detailed labels of the parkland scale and its components are shown in TABLE \ref{tab:pgs_dataset}, followed by their definition in \cite{madni2018parkland}. Videos are randomly split into a $90/10$ ratio for training/testing purposes. 3 random splits were used for the experiments.

\subsubsection{CVS100} CVS 100 is composed of a total of videos. The dataset was annotated by clinical experts with regard to the CVS. Within the dataset, in a total of 50 videos CVS was achieved whereas in 50 videos CVS was not achieved. The videos were down-sampled to 1 fps. A total of 50 frames were extracted and annotated prior to 'checkpoint 1', marking the time point before application of the first clip to either cystic duct or the artery, which can be regarded as the 'point of no return' as CVS achievement is unobtainable beyond that point \cite{ban2021aggregating}. Labels included all components of CVS as defined by the Society of American Gastrointestinal and Endoscopic Surgeons, including $\{$~\textit{cvs, cystic artery, cystic duct, two and only two ductal structures leading to the gallbladder, cystic plate being dissected and visibility of the liver between the two ductal structures} $\}$ \cite{strasberg2010rationale, mascagni2021artificial}. The training/testing ratio was also set to $90/10$.

\begin{figure*}[th!]
\vspace{1ex}
\centering
\begin{tabular}{c}
\includegraphics[width=0.98\textwidth]{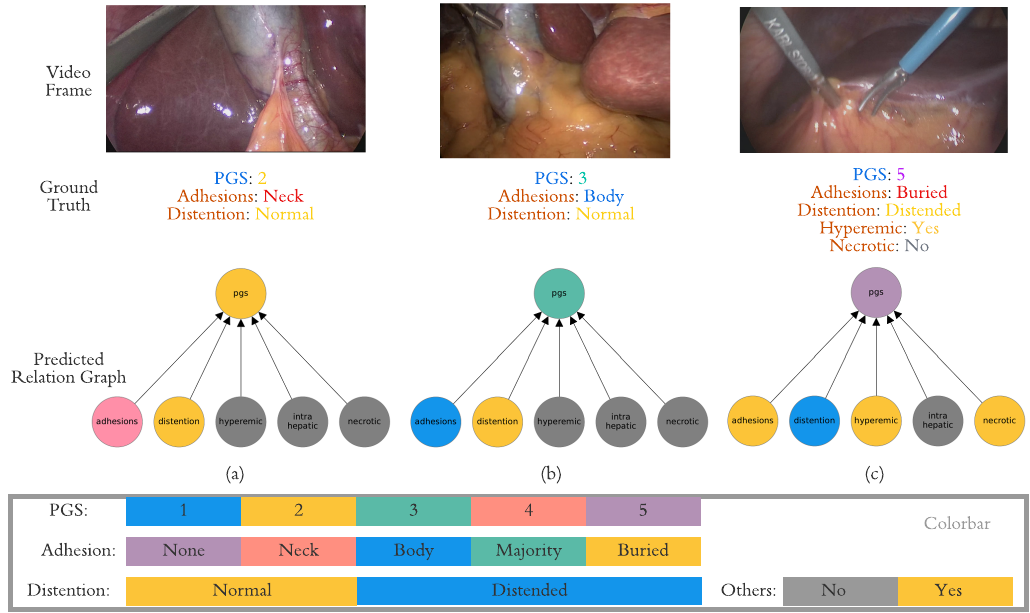}
\end{tabular}
\caption{\color{green}{ Example results on PGS200 dataset. First row: the last frame of the sequence. Second row: indicates the ground truth activations. Third row: the estimated relation graph. different colors indicate different activation level.}}
\label{fig:results_pgs}
\end{figure*}

\begin{table}[th!]
    \caption{Results on PGS200 Dataset. We note the improvement in performance when leveraging temporal information in CNN-LSTM, Temporal MTL approaches. Our ConceptNet further improves upon such approaches by leveraging the problem's reasoning structure.}
    \centering
    \begin{tabular}{c|c|c}
    \hline
    Method & Accuracy ($\uparrow$) & Average Distance ($\downarrow$) \\
    \hline
    MTL \cite{nwoye2020recognition} (MICCAI 20)& 37.0 $\pm$  17.2 & 1.21 $\pm$ 0.4 \\
    Tripnet \cite{nwoye2020recognition}(MICCAI 20)& 27.3 $\pm$ 4.5 & 1.35 $\pm$ 0.18 \\
    Resnet50-LSTM & 44.5 $\pm$ 12.8 &  0.72 $\pm$ 0.17\\
    I3D \cite{carreira2017quo} (CVPR 17) &  53.8 $\pm$ 15.5 & 0.57 $\pm$ 0.2\\
    CSN \cite{pan2019cross} (CVPRW 19) & 53.7 $\pm$  14.8 & 0.58 $\pm$ 0.18\\
    TSM \cite{lin2019tsm} (ICCV 19)& 62.1 $\pm$ 29.6 &  0.54 $\pm$ 0.43\\
    \hline
    \textbf{ConceptNet-Resnet50 (Ours)}  & 54.6 $\pm$ 21.3 &  0.58 $\pm$ 0.08\\
    \textbf{ConceptNet-VIT (Ours)}  & \textbf{67.5} $\pm$ \textbf{28.2} &  \textbf{0.46} $\pm$ \textbf{0.4}\\
    \hline
    \end{tabular}
    \label{tab:pgs}
\end{table}

\subsection{Settings}
\subsubsection{Training Settings} Within the concept model, the dimension of the LSTM was set to 128 and the length of the LSTM sequence to 8. During training, the learning rate was $2 \times 10^{-4}$, and the model was trained for 30, 20, 10 epochs for the Action triplet, PGS and CVS tasks separately. The focal loss with ($\gamma =2$) was applied to all three problems.  
\vspace{-1ex}
\subsubsection{Visual encoder} ResNet50 was adopted as a visual encoder baseline, with the last classification layer removed for feature extraction. For Vision Transformer encoder, VIT-base model was used, where the patch size was 16, and only attention blocks 5-11 were tuned to train different tasks. Both ResNet and VIT were pre-trained on Imagenet dataset. 
\subsubsection{Ordinal classification} Since the PGS is a discrete learning problem of the relevant order, we converted rank estimation to an ordinal classification problem\cite{cheng2008neural}. That means we convert onehot vectors to ordinal vectors for classification (e.g. PGS = 3, onehot vector: $[0,0,1,0,0]$, ordinal vector: $[1,1,1,0,0]$). Performing ordinal classification allows us to presume a better structure of the PGS problem.

\subsection{Evaluation Metric}
We use different metrics to evaluate the performance of the proposed model. \\
\noindent \textbf{Average Precision (AP)} We use AP to measure the accuracy of the estimates, which is defined as the area under the precision-recall
curve. Different AP statistics are used to evaluate the joint activation of the components (e.g. $AP_{IVT}$ for joint activation of Instrument-Verb-Target). In practice the toolbox \footnote{https://github.com/CAMMA-public/ivtmetrics} provided by \cite{nwoye2022data} are used to calculate the APs.
\noindent \textbf{Accuracy} The first metric is the accuracy for the multi-class classification performance (e.g. PGS), which is defined as $(TP + TN) / (TP + TN + FP + FN)$.\\
\noindent \textbf{Balanced accuracy (BA)} is often used to evaluate the binary classifier and has proven particularly useful for imbalanced classes. It is defined as $BA = 0.5 \times (TP/(TP + FN)) + 0.5 \times (TN/(TN + FP))$. \\ 
\noindent \textbf{Average Distance (AD)} In addition to accuracy, we use AD to measure the performance of the algorithm on the PGS, which is the $L_1$ distance between the ground truth and the estimation. \\

\begin{figure*}[t!]
\vspace{-0ex}
\centering
\begin{tabular}{c}
\includegraphics[width=1.00\textwidth]{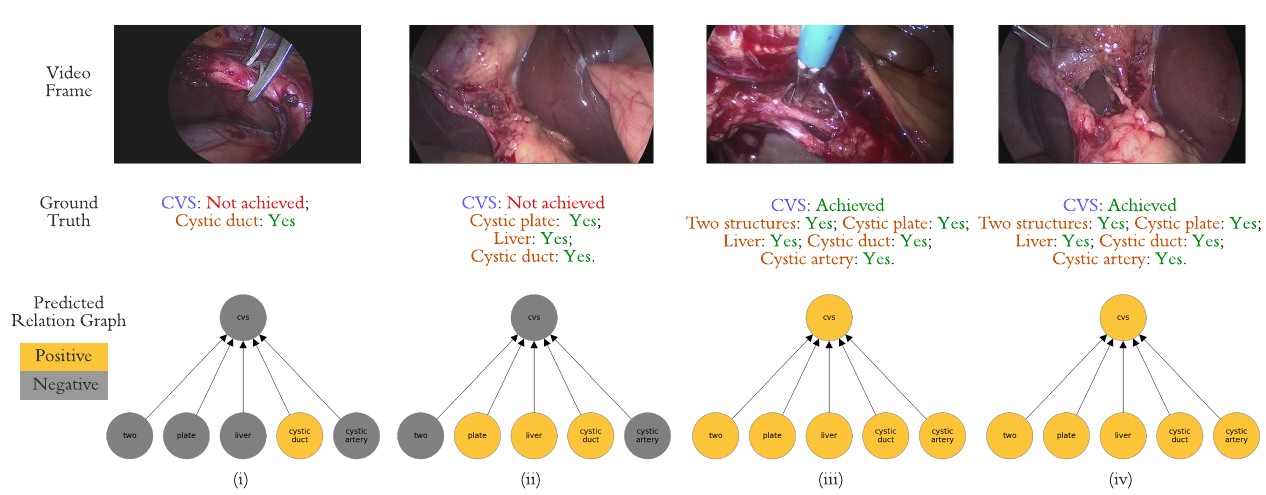}
\end{tabular}
\caption{\color{green}{Example results on CVS100 dataset. (a) is the last frame of the sequence. (b) indicates the ground truth activations. (c) is the estimated graph activation. (i) and (ii) are two examples when CVS is partially achieved/ (iii) and (iv) are two examples when CVS is fully achieved.}}
\label{fig:results_cvs}
\vspace{-2ex}
\end{figure*}


\subsection{ActionTriplet Recognition}

A set of examples are shown in Fig. \ref{fig:results_triplet}, where in column (a) the last frame is perceived by the network. Columns (b) show the ground-truth activation of the instrument, verb, tissue with round nodes, and the link represents the joint triplet interaction. Column (c) shows the estimation with the same legends. In example (i), the model correctly estimated the activation of the nodes "grasper", "grasp" and "specimen bag", shown as round nodes. Also, the joint activation of "grasper - grasp - specimen bag" is activated, shown as the green arrows. Example (ii) displays a more difficult case, since two different tools are present in the video and performing different actions. In addition to correctly estimating the nodes. The model correctly predicts the triplets "hook - dissect - gallbladder" and "grasper - retract - gallbladder" the same time. Quantitative results are shown in Table \ref{tab:results_triplet}. 
As commonly done for this problem \cite{nwoye2023cholectriplet2021}, in order to capture the highly nonlinear interactions between the concepts, we compute the action triplet as a combination of the per-edge computation and the product of the emitted probabilities of the associated nodes, following the 
$
  P_{trip} =  0.5 * P_{trip}^{H} + 0.5 * (P_{inst}^{N} * P_{verb}^{N} * P_{target}^{N}) ^{1/3}  
$
where $P_{\cdot}^{N}$ denotes the probabilities per node. This sets a convenient working point for edge emissions as the residual of $P_{trip}$ and improves network performance. 
\begin{table*}[t!]
    \caption{Average Accuracy for CVS100 Dataset. As can be seen, our approach leads to \\superior results w.r.t existing multitask and structured approaches.}
    \centering
    \begin{tabular}{c|c|c|c|c|c|c}
    \hline
    \multirow{ 2}{*}{Method} & \multirow{ 2}{*}{CVS} & \multicolumn{5}{c}{Individual Components}\\
    \cline{3-7}
    && Cystic artery & Cystic Duct & Liver & Cystic Plate & Two structures
    \\
    \hline

    MTL \cite{nwoye2020recognition} (MICCAI 20) & 52.3\% &  59.4\% &  55.7 \% & 50.8 \%&  48.6\% & 49.5\%\\
    Tripnet \cite{nwoye2020recognition} (MICCAI 20) & 50.1\% &  49.9\% &  49.9\% & 49.8\% & 50.2\% & 50.0\%  \\
    Resnet50-LSTM & 55.3 \% &  - & - & - &  - & - \\
    I3D \cite{carreira2017quo} (CVPR 17) & 49.9\% &  45.1 \% & 50.0 \% & 49.4 \% & 47.1 \% & 49.1 \% \\
    CSN \cite{pan2019cross} (CVPRW 19) & 50.0\% & 47.8 \% & 50.5\% & 52.5 \% & 53.3\% & 49.9 \% \\
    TSM \cite{lin2019tsm} (ICCV 19)& 57.7\% & 61.0 \% & 59.5\% & 61.2\% & 63.4 \% & 55.6 \% \\
    \hline
    \textbf{ConceptNet-Resnet50 (Ours)}  & 62.4 \% &  55.0 \% & 53.8 \% &  60.1 \%  & 58.4 \% & 52.5 \% \\
    \textbf{ConceptNet-VIT (Ours)}  &\textbf{67.1 \%}& 55.1\% & 55.2 \% & 60.2 \% & 55.6 \% & 53.9 \%\\
    \hline
    \end{tabular}

    \label{tab:cvs}
\end{table*}
\addnote[R2Q4b]{1}{
Comparing our approach to other approaches used on the dataset, including both attention-based Tripnet and multi-head attention transformer Rendezvous\cite{nwoye2022rendezvous} network approaches. Our approach works harmoniously  with attention-based visual encoders \cite{dosovitskiy2020image}, as we see in the \texttt{ConceptNet-VIT} row of table~\ref{tab:results_triplet}. Thanks to the proposed graph design, the message passing focuses on known triplet relations, resulting in more accurate estimation results, with the best instrument-verb-tissue average precision $AP_{IVT}$ of $30.6 \pm 1.9\%$. 
We note that our approach surpasses the other methods,  both  heavy attention mechanisms or other task-specific adaptations. The performance advantage of our model is because the concept message-passing constrains the required visual encoding tasks into loosely defined semantics needed per frame, and limits the spatiotemporal processing to an efficient pass, whereas the visual transformers excel at the extraction of image-specific cues for these semantics, resulting in a semi-structured attention mechanism across the processing of the video data.
}
\subsection{Parkland Grading Scale}

One application of the model we demonstrate is the estimation of PGS. The qualitative results can be found in Fig. \ref{fig:results_pgs}. All three example results in the figure demonstrate, that the estimation of the graph activation \ref{fig:results_pgs}(c) is close to the ground truth \ref{fig:results_pgs}(b). Notably, the proposed model is able to estimate all components of the PGS accurately. By correctly estimating the components, the model is also able to aggregate the node information and hence achieve the correct estimation on the overall PGS. Fig. \ref{fig:results_pgs} shows three examples of different stages of gallbladder inflammation. (i) is a normal gallbladder of PGS 1, with a mere adhesion to the neck of the gallbladder and a normal degree of distention.  (iii) shows a pathologically inflamed gallbladder. The model accurately estimated the PGS to be 5 in the latter, by correctly identifying major adhesions obstructing the view of the gallbladder. We also observed that the model accurately estimates the presence of necrosis, which is always associated with major inflammation. However, the structured network incorrectly concludes that if PGS 5 is detected, then necrosis should also be activated. 

The quantitative results are also shown in Table \ref{tab:pgs}. We compared the model with two sets of baselines. The first set of the baselines are several existing models for surgical video understanding, namely Resnet50-LSTM, MTL and Tripnet. CNN-LSTM uses a Resnet\cite{he2016deep} image encoder, followed by an LSTM, which is a simplified version of the ConceptNet-ResNet50 by removing the graph message passing. By Multitask learning (MTL) and Tripnet we refer to two adaptations of \cite{nwoye2020recognition} to address the PGS recognition task. MTL uses the image as input, then provides the emission for each component of PGS. An additional layer is used to fuse all component probabilities to infer the overall PGS. In contrast to MTL, Tripnet projects all the probabilities of the components into a joint feature space. The PGS scales correspond then to the different positions in the joint space, as demonstrated by \cite{madni2018parkland}. For example, PGS 1 is represented by the following in the joint space $ \textit{adhesion:none}$, $ \textit{distention:normal}$, $ \textit{hyperemic:no}$, $ \textit{intra-hepatic:no}$, $ \textit{necrotic:no}$. 
\color{green}
\noindent The second set of baselines are SOTA methods in the computer vision action recognition field, which include I3D, CSN and TSM.
\color{black}

\addnote[R2Q4c]{1}{
Among the baselines, Tripnet is a structured approach that encodes the interaction information in a joint space. However, encoding the interaction by simply multiplying the emissions for each component may not be suitable for PGS recognition since the individual concepts are abstract, which yields a low performance in PGS recognition task. Aside from Triplet, other spatial-temporal networks lack the problem structure information, and when under limited data conditions they suffer from achieving the best results.   
In contrast to these existing baseline models, the proposed ConceptNet involves both spatial-temporal modeling and the specific problem relation encoding in the network. The ConceptNet-VIT achieved the highest classification average accuracy of $67.5\%$, and the lowest average distance of $0.46$. Hence our model demonstrates significantly better results in the performed experiments. }\\

\vspace{-3ex}
\subsection{Critical View of Safety}

In the following section we show the results of the proposed model on the inference of the CVS. During the temporal evolution of CVS, the proposed model can correctly estimate the activation of each individual component as described in \ref{sec:dataset}. In Fig. \ref{fig:results_cvs} we show a) various frames of laparoscopic cholecystectomy around the time point of CVS achievement, b) the annotated ground truth and c) the model estimation. The activations of the nodes are illustrated in yellow. The figure clearly illustrates an accurate estimation despite partially difficult viewpoints and optical variations. The annotation labels address all individual components of CVS, which are achieved in a temporally consecutive order. \addnote[R2Q4d]{1}{The quantitative results are illustrated in Table~\ref{tab:cvs}. As shown in Table~\ref{tab:cvs}, both versions of the proposed approach have achieved superior results to existing approaches on CVS detection. with $62.4\%$ for ConceptNet-ResNet50 and $67.1\%$ for ConceptNet-VIT. Also, the  labels that relate to its individual components can be used to leverage the additional supervision to improve performance. The structure of the knowledge graph within the proposed ConceptNet allows for the detection of other relevant points around the crucial time point of clipping and cutting the cystic duct and artery.}

\label{sec:results phase}

\section{Conclusion}
We demonstrate a new, general framework for temporal analysis of surgical video data, which affords easy embedding of surgical concepts via a knowledge-graph structured network. The framework demonstrates superior results on several important surgical applications compared to existing approaches. As the framework lends itself to a gradual expansion of the knowledge graph, we intend to augment the graph with additional applications and concepts, accounting for various notions of surgical safety. This presents a significant step towards a bigger network that can establish a collective computational surgical consciousness\cite{hashimoto2018artificial} and generalize over different surgical procedures and ML tasks.

\bibliographystyle{ieeetr}
\bibliography{references}
\end{document}